\documentclass{article}

\usepackage{microtype}
\usepackage{graphicx}
\usepackage{subfigure}
\usepackage{booktabs} 

\usepackage{hyperref}

\usepackage[accepted]{icml2023}

\usepackage{amsmath}
\usepackage{amssymb}
\usepackage{mathtools}
\usepackage{amsthm}

\usepackage[capitalize,noabbrev]{cleveref}

\theoremstyle{plain}

\theoremstyle{definition}

\theoremstyle{remark}

\usepackage[textsize=tiny]{todonotes}

\icmltitlerunning{Submission and Formatting Instructions for ICML 2023}

\begin{document}

\twocolumn[
\icmltitle{OC-NMN: Object-centric Compositional Neural Module Network for Generative Visual Analogical Reasoning}



\icmlsetsymbol{equal}{*}

\begin{icmlauthorlist}
\icmlauthor{Rim Assouel}{yyy,s}
\icmlauthor{Pau Rodriguez}{s}
\icmlauthor{Perouz Taslakian }{s}
\icmlauthor{David Vazquez }{s}
\icmlauthor{Yoshua Bengio }{yyy}

\end{icmlauthorlist}

\icmlaffiliation{yyy}{Mila}
\icmlaffiliation{s}{ServiceNow Research}

\icmlcorrespondingauthor{Rim Assouel}{assouelr@mila.quebec}

\icmlkeywords{Machine Learning, ICML}

\vskip 0.3in
]



\printAffiliationsAndNotice{\icmlEqualContribution} 

\begin{abstract}
A key aspect of human intelligence is the ability to imagine --- composing learned concepts in novel ways --- to make sense of new scenarios. Such capacity is not yet attained for machine learning systems. In this work, in the context of visual reasoning, we show how modularity can be leveraged to derive a compositional data augmentation framework inspired by imagination. Our method, denoted \textbf{O}bject-centric \textbf{C}ompositonal \textbf{N}eural \textbf{M}odule \textbf{N}etwork (OC-NMN), decomposes visual generative reasoning tasks into a series of primitives applied to objects without using a domain-specific language. We show that our modular architectural choices can be used to generate new training tasks that lead to better out-of-distribution generalization. We compare our model to existing and new baselines in proposed visual reasoning benchmark that consists of applying arithmetic operations to MNIST digits.
\end{abstract}

\section{Introduction}
Humans have the remarkable ability to adapt to new unseen environments with little experience~\citep{lake2017building}. In contrast, machine learning systems are sensitive to distribution shifts~\citep{arjovsky2019invariant,su2019one,engstrom2019exploring}. One of the key aspects that makes human learning so robust is the ability to produce or acquire new knowledge by composing a few learned concepts in novel ways, an ability known as compositional generalization~\citep{fodor1988connectionism,lake2017building}. 
The question of how to achieve such compositional generalization in both humans and machines remains an active area of research~\citep{ruis2022improving}. 


Both imagination and abstraction are core to human intelligence. Objects, in particular, are an important representation used by the human brain when applying analogical reasoning~\citep{spelke2000core}. For instance, we can infer the properties of a new object by transferring our knowledge of these properties from similar objects~\citep{mitchell2021abstraction}. This realization has inspired a recent body of work that focuses on learning models that discover objects in a visual scene without supervision~\citep{air, sqair, nem, rnem, iodine, monet, objects, locatello2020objectcentric}. Many of these works propose inductive biases that lead to a decomposition of the visual scene in terms of its constituting objects. The expectation is that such an  object-centric decomposition would lead to better generalization since it better represents the underlying structure of the physical world~\citep{parascandolo2018learning}. 

Most visual reasoning benchmarks revolve around variations of Raven's Progressive Matrices (RPM)~\citep{james1936raven,zhang2019raven,barrett2018measuring,hoshen2017iq}, all of which are \textit{discriminative} tasks in which the solver chooses from a set of candidate answers.
However, in a recent survey, \citet{mitchell2021abstraction} argues that models trained on discriminative tasks are prone to shortcut learning and thus recommends evaluating models on generative tasks that focus on human core knowledge~\citep{spelke2000core}.  \citet{mitchell2021abstraction} also argues that systems that generate answers are in many cases more interpretable. To that end, \citet{chollet2019on} proposed a generative reasoning task, the Abstract Reasoning Corpus (ARC), where the model is given a few examples of input-output (I/O) pairs and has to understand the underlying common program that was applied to the inputs to obtain the outputs. ARC tasks rely only on the innate core knowledge systems, including intuitive knowledge about objects, agents and their goals, numerosity, and basic spatial-temporal concepts. However, ARC  remains a challenging task unapproachable by current deep learning methods. In this work, we  take a step towards addressing the ARC challenge by designing a new and simpler generative benchmark, which we call Arith-MNIST. Using this benchmark, we evaluate the systematic compositional generalization of models on a set of controlled and easily extendable axes and show the benefits of object-centric inductive biases.

To tackle Arith-MNIST, we propose the \textbf{O}bject-centric \textbf{C}ompositional- \textbf{N}eural \textbf{M}odule \textbf{N}etwork (OC-NMN), an example of how object-centric inductive biases can be exploited to  (1) design a modular  architecture that can solve  generative visual reasoning tasks like ARC, and (2) derive a compositional data augmentation paradigm that we lead to better out-of-distribution generalization.    
The core idea underlying OC-NMN is to predict a \textit{neural template}~\citep{reed2015neural, cai2017making, li2020closed} that specifies a task-specific composition of neural modules. The neural modules can be reassembled in unseen ways to invent new tasks.  

Our contribution is as follows: 

\begin{itemize}
    \item We propose a benchmark Arith-MNIST that serves as a test-bed to evaluate compositional generalization capabilities of both visual analogical reasoning models and object-centric perception models. The benchmark is constructed around a set of controllable primitives (e.g., arithmetic operations over visual digits) that can be easily extended.
    \item We propose the model OC-NMN that adapts neural module networks (NMN)~\citep{andreas2016neural} to solve visual generative analogical reasoning tasks.
    \item We show how such modular inductive biases can be leveraged to derive a compositional imagination framework. We show that samples created within this framework help with systematic generalization.
    \item Finally, we show that the ability of the perception part of the models to represent and disentangle object-level attributes is key to generalizing to unseen combinations of attributes.
\end{itemize}

\section{Related Work}
\paragraph{Object-centric Representation.} A recent research direction explores unsupervised object-centric representation learning from visual inputs ~\citep{locatello2020objectcentric, monet, iodine,eslami2016attend, crawford2019spatially,stelzner2019faster,lin2020improving, geirhos2018imagenettrained}. The main motivation behind this line of work is to disentangle a latent representation in terms of objects composing the visual scene (e.g., slots). Recent approaches to slot-based representation learning focus on the generative abilities of the models; in our case, we study the impact of object-centric inductive biases on the systematic generalization of models in a visual reasoning task. We observe that the modularity of representations is as important as the mechanisms that operate on them~\citep{goyal2020object,goyal2021neural}. Additionally, we show that object-centric inductive biases of both representations and mechanisms allow us to derive a compositional imagination framework that leads to better systematic generalization.
\paragraph{Modularity.} Extensive work from the cognitive neuroscience literature \citep{Baars1997INTT,deahene} suggests that the human brain represents knowledge in a modular way, with different parts (e.g., modules) interacting with a working memory bottleneck via attention mechanisms. Following these observations, a line of work in machine learning \cite{biases2020,goyal2019recurrent, goyal2020object, goyal2021neural,ostapenko2021continual,goyal2022inductive} has proposed to translate these characteristics into architectural inductive biases for deep neural networks.  Recent approaches have explored architectures composed of independently parameterized modules competing with each other to communicate and attend or process the input~\citep{goyal2019recurrent, goyal2020object,goyal2021neural}. Such architectures are inspired by the notion of independent mechanisms~\citep{pearl2009, bengio2019meta,goyal2019recurrent,goyal2022inductive}, which suggests that a set of independently parameterized modules capturing causal mechanisms should remain robust to distribution shifts caused by interventions, as adapting one module should not require adapting the others. The hope is that  out-of-distribution (OOD) generalization would be facilitated by making it possible to sequentially compose the computations performed by these modules, whereby novel combinations of existing concepts can explain new situations. Neural Module Networks (NMNs) \citep{andreas2016neural} have also been shown to be a promising approach for systematic generalization in the context of Visual Question Answering (VQA) \citep{closure,sysgen}. Their main inductive bias is selecting a question-specific neural network layout containing a sequence of sub-tasks where each sub-task corresponds to a module.  In this work, we adapt NMNs main inductive bias to solve generative visual analogical reasoning tasks. We also show how our modular architectural choices can be exploited to derive a compositional data augmentation framework that allows better compositional generalization; we do so by explicitly exposing the model to data samples composed of novel combination of learned concepts.

\paragraph{Imagination, Dreaming, and Generalization}
Dreams are a form of imagination that has inspired significant influential work~\citep{Hinton06,ellis2021dreamcoder,hafner2019dream,hafner2020mastering}. An interesting explanation for such a phenomenon is the \textit{overfitted brain hypothesis} (OBH)~\citep{hoel2021overfitted}, which states that dreaming improves the  generalization and robustness of learned representations. The idea is that, while dreaming, the brain recombines patterns seen during wake time. This  results in artificial data augmentation in the form of dreams that regularizes the brain and prevent overfitting the patterns seen while awake.

Dreamcoder~\citep{ellis2021dreamcoder} is a recent example where training on imagined patterns improves generalization for program induction. Program induction is a challenging problem because the search space is combinatorially large, and new unseen programs have low likelihood. To address these challenges, Dreamcoder leverages a wake-sleep algorithm that reduces the search space by learning a domain-specific language (DSL) while learning to search programs efficiently. 
 During training, Dreamcoder undergoes a \textit{dreaming} phase where the model learns to solve new tasks generated by sampling programs from a DSL and applying them to inputs seen during the \textit{wake} phase. Although Dreamcoder is promising for program induction, the DSL is a major roadblock to solving open-ended visual reasoning tasks where the input consists of raw pixels rather than symbols. In this work, we show promising results on overcoming these challenges by relying on object-centric inductive biases.
\section{Arith-MNIST: Generative Arithmetic Visual Analogical Reasoning}\label{section:dataset}

\begin{figure*}[ht!]
    \centering
    \includegraphics[trim={35 40 35 35},clip,width=0.9\textwidth]{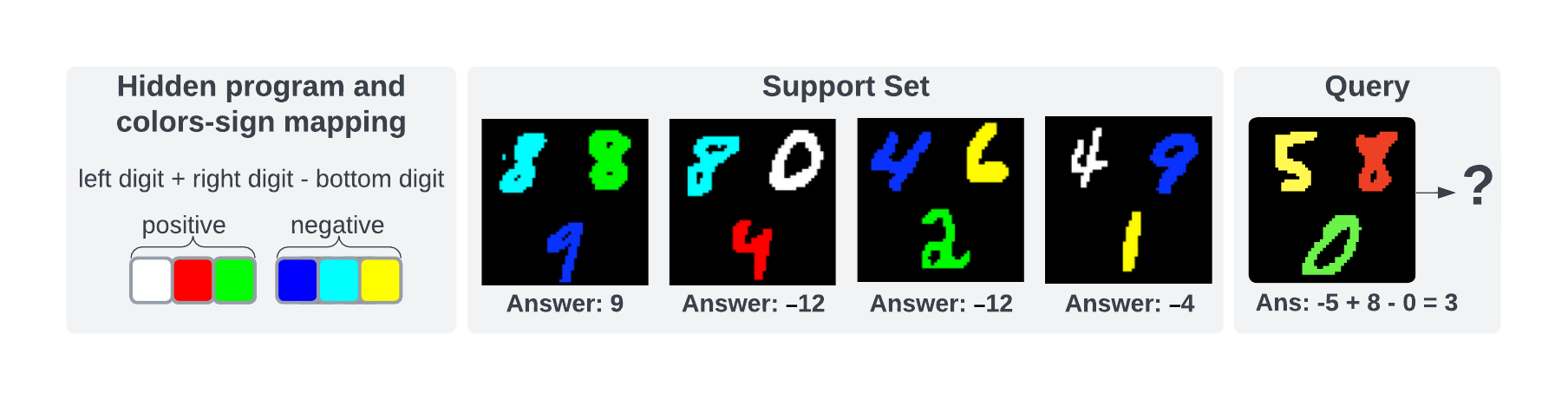}
    \caption{\textbf{Arith-MNIST data sample}. The color/sign mapping is given on the left. The model must also infer the hidden program to generate the right output for the query.}
    \label{fig : data}
\end{figure*}
Arith-MNIST is made of visual generative analogical reasoning tasks. In generative analogical reasoning tasks such as the ones in ARC~\citep{chollet2019on}, a model is presented with a few input-output examples (the \emph{support set}) and is asked to predict the output of new input queries of the same task (the \emph{query set}). The outputs from the support set are obtained by applying a program (the same for all the elements of one support set)  to the inputs. To predict the right outputs of the input query set, a model needs to infer the program applied to the support and apply this program to the queries. 
Given that current ML systems still struggle to tackle ARC,  we propose a simpler set of tasks that involve a set of controlled arithmetic operations on digits. 



In our benchmark, inputs  are $56 \times 56$ images with three colored MNIST digits (unsigned) placed at three different positions. These visual digits can have values between $-9$ and $9$ and their color represents their sign. There are six different colors in total ($3$ of them are \textit{negative} and the remaining $3$ are \textit{positive}). 
The program applied to the inputs is a sequence of arithmetic operations in a specified order. The order in which we select the digits is given by a sequence of conditions (e.g. position in the image, maximum digit, etc\dots). We give those primitives in \Cref{table : split} and more details in the Appendix. 

The benchmark is split into different datasets that are composed of a varying number of tasks and examples per task. These split and their composition is given in \Cref{table : split}. The hard split approaches the ARC setting the most since only a handful of examples per tasks are in the training set.  Since we want to measure models' compositional generalization we create different sub-splits that aim at evaluating different axes of compositional generalization. We give more details about the splits and the sub-splits in \Cref{sub : exp} and the Appendix. We give an example of the benchmark in \Cref{fig : data}.

\begin{table}[h!]\label{tab : data}
\renewcommand{\arraystretch}{1.1}
\centering
 \resizebox{\linewidth}{!}{\begin{tabular}{||c || c c| c  ||} 
 \hline
 Dataset & \multicolumn{2}{c|}{Primitives} & $\#$ of Training Tasks \\ 
 \hline\hline
   Split & Conditions & Operations  &   \\
  \hline
 Easy & \texttt{position} & \texttt{add, sub} & 5 \\ 
 Medium & \texttt{position, max, min} & \texttt{add, sub, or, xor, cat, invcat} & 150 \\
 Hard & \texttt{position, max, min} & \texttt{add, sub, or, xor, cat, invcat} & 1980 \\
 Percep & \texttt{position} & \texttt{add, sub} & 5 \\ 
 \hline
 \end{tabular}}
 \vskip0.2cm
 \caption{Dataset Splits Description.}
 \label{table : split}
\end{table}
\paragraph{Text mode.} We provide users the option to skip the perception part (understanding the images) by providing a textual version of our dataset. This enables evaluating compositional generalization reasoning capabilities of existing large language models (LLMs). 
In this version, each image/output pair is described in natural language text (e.g., \texttt{There is a blue four on the left, a red five on the right, and a green seven at the bottom. The result is 4}). 
 
\section{Object-centric Compositional Neural Module Network}

Our model \textit{OC-NMN} (\textbf{O}bject-centric \textbf{C}ompositional \textbf{N}eural \textbf{M}odule \textbf{N}etwork) can be seen as part of the  neural module networks (NMNs) family of models, adapted to generative visual analogical reasoning tasks like ARC. NMNs are mainly used in the context of visual question answering and assemble a question-specific composition of network modules, a \emph{neural template}. This neural template can further be applied to different input images to answer the same question.  NMNs typically consist of two main components: the \emph{controller module} and the \emph{execution module}. 
The controller module processes the input question and generates a neural template that, when executed, generates the answer to the question. This neural template specifies the sequence of modules to be executed along with their arguments (coming from the question). 
The execution module takes the neural template predicted by the controller, along with the input image, and generates the answer by assembling the modules specified by the template. 

\begin{figure}[h!]
    \centering
    \includegraphics[trim={0 0 0 0},clip,width=\linewidth]{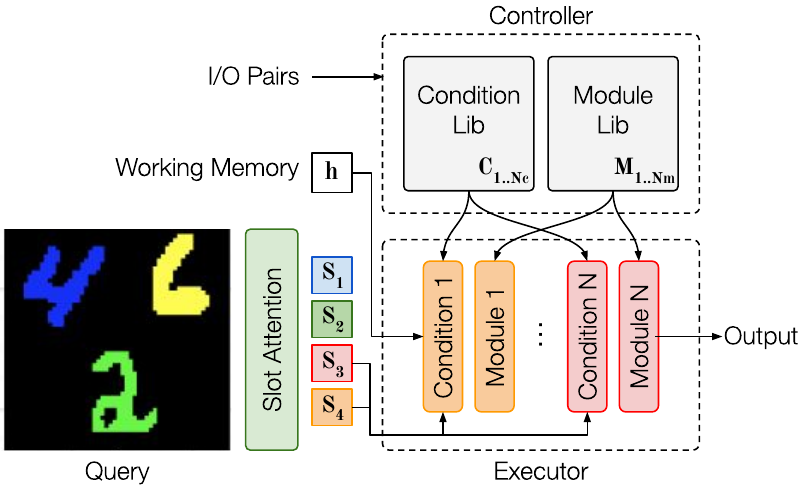}
    \caption{\textbf{Model overview}. Based on I/O pairs, the controller selects a series of conditions and modules, forming a neural program executed on the slots, producing an output for a query.}
    \label{fig:model}
\end{figure}

Following this general structure of NMNs, \textit{OC-NMN} also predicts a sequence of modules to be assembled along with their arguments; however, our module generation differs from that of NMNs in two ways: (1) the model does not have access to the program (e.g. the parsed question in the VQA setting) and needs to infer it from the demonstration pairs in the support set,  and, (2) the arguments to the modules come directly from the visual input query. Our model overview is given in \Cref{fig:model}

The computation steps (and parts of OC-NMN) are the following : (1) A \textbf{perception network} maps the visual inputs $\mathbf{x}$  to $N_s$ object-centric slots $\mathbf{S} = [\mathbf{S}_1, \dots,\mathbf{S}_{N_s}]$ using a slot attention~\citep{locatello2020objectcentric} mechanism  then,
(2) a \textbf{controller} takes the support set as input and outputs a task embedding $\mathbf{z}$. (3) A \textbf{selection bottleneck} translates this task embedding into a sequential neural template; finally, (4) the \textbf{executor} takes this neural network template along with an input query (i.e., its object-centric slots) and performs the sequential updates. The controller module's architecture is the same for all the baselines considered (including our model) and corresponds to the  Differentiable Neural Computer controller \citep{dnc} proposed in Neural Abstract Reasoner~\citep{nar}. Our contribution resides in the design of the executor and the Selection Bottleneck that we detail below. 

\subsection{Executor}
The executor takes a visual input query $\mathbf{x} \in \mathbb{R}^{56 \times 56 \times 3}$ and a neural template. The execution consists of sequentially transforming a working memory $\mathbf{h}$: at each time-step $t$. The neural template specifies a module (from a learnt modules library) that will transform the working memory $\mathbf{h}$ and a condition (from a learnt conditions library) that will select an argument from the visual input query $\mathbf{x}$.

The visual input is first mapped to a set of $N_s$ object-centric slots $[\mathbf{S}_1, .., \mathbf{S}_{N_s}]$ that is later used as candidate arguments for each update of the working memory. 
The executor is composed of a library of $N_r$ learned  modules (e.g. rules, implemented as small GRU cells) and $N_c$ condition values. The conditions are expected to encode how to select an argument (for instance: select the highest digit among slots) for the update. Both modules and conditions are indexed by  learned tags $\mathbf{M} = [\mathbf{M}_1,\dots, \mathbf{M}_{N_r}]$ and $\mathbf{C} = [\mathbf{C}_1, \dots, \mathbf{C}_{N_c}]$.

The neural template comes from the selection bottleneck. It specifies: 
\begin{itemize}
    \item the number  of updates $T$ the executor needs to perform. A sequence of scalar gates specifies it at each time step, which we denote by  $g = [g_1, \dots,g_T]$;
    \item the sequence of modules  
    that will perform the $T$ updates of the working memory, which we denote [$\hat{m}_1$, \dots, $\hat{m}_T$]; 
    \item the sequence of conditions that select the slots used as arguments, denoted [$\hat{c}_1$, \dots, $\hat{c}_T$].
\end{itemize}  

\paragraph{Argument Selection.}
At each time step, a slot argument is selected through a key-query attention mechanism. At time-step $t$, the condition vector $\hat{c}_t$ is compared against all the input slots to select the one that corresponds best to the features encoded in the condition (e.g., select the slot at the ``top-left'' of the image). 
In the attention mechanism, the query comes from  the condition vector $\hat{c}_t \in \mathbb{R}^{1 \times d}$ and the keys come from the  $N_s$ slots $\mathbf{S} = [\mathbf{S}_1,\dots, \mathbf{S}_{N_s} ] \in \mathbb{R}^{N_s \times d}$ (transformed by a self-attention module to extract the keys).The selected slot argument $\mathbf{\hat{s}}_t$ at time-step $t$ is given by: 
\begin{equation}   \label{eq:argument_selection}
    \mathbf{\hat{s}}_t = \text{GumbelSoftmax}(\frac{\hat{c}_t\mathbf{S}^T}{\sqrt{d}})\mathbf{S}  \in \mathbb{R}^{1 \times d}
\end{equation}

\paragraph{Sequential Update.} Given a sequence of  modules [$\hat{m}_1$, \dots, $\hat{m}_T$], a sequence of input arguments [$\mathbf{\hat{s}}_0$, \dots, $\hat{s}_T$] and a sequence of gates [$g_1$, ..., $g_T$], the executor updates a working memory whose state at time step $t$ is denoted by $\mathbf{h}_t$ such that: 
\begin{equation}
\label{eq:state_update}
   \mathbf{h}_{t+1} = (g_{t+1}) \cdot \mathbf{h}_{t} + (1 - g_{t+1} )\cdot\mathbf{\hat{m}}_t(\mathbf{\hat{s}}_{t+1}, \mathbf{h}_t) \text{~and~} \mathbf{h}_0 = \mathbf{\hat{s}}_0
\end{equation}
For ease of notation, we let $\textbf{executor}(\mathbf{x}, \mathcal{P})$ be the result of applying the neural template $\mathcal{P}$ to $\mathbf{x}$.
\subsection{Selection Bottleneck}
We describe here the component that translates the task embedding $\mathbf{z} = \{\mathbf{z}_1, .., \mathbf{z}_T \}$ from the controller to a neural template that the execution module will use. At each time-step $t$ we need to predict a module (from a learnt modules library)  and a condition vector (from a learnt conditions library). We propose to predict those with a key-query attention mechanism: at each time-step $t$, the task embedding $\mathbf{z}_t$ is compared to the $N_r$ learned module tags $\mathbf{M}$  and the $N_c$ learned conditions tags $\mathbf{C}$. The keys are extracted from the condition/module tags, whereas the query is extracted from the task embedding  $\mathbf{z}_t$ (using two MLPs $Q_r$ and $Q_c$) such that the $t$-th element of each sequence is obtained with:
\begin{equation}
    W_m^t = \text{GumbelSoftmax}(\frac{Q_r(\mathbf{z_t})\mathbf{M}^T}{\sqrt{d}}) \in \mathbb{R}^{1 \times N_r}
\end{equation}
and the resulting update is given by the following weighted sum :
\begin{equation}
    \hat{m}_t(\mathbf{h}_t, \mathbf{\hat{s}}_t) = \sum_{i = 1}^{N_r} W_m^t[i] \cdot m_i(\mathbf{h}_t, \mathbf{\hat{s}}_t).
\end{equation}
Similarly, the conditions are obtained through: 
\begin{equation}
    \hat{\mathbf{c}}_t = \text{GumbelSoftmax}(\frac{Q_c(\mathbf{z_t})\mathbf{C}^T}{\sqrt{d}})\mathbf{c} 
\end{equation}
with $\mathbf{c} \in \mathbb{R}^{N_c \times d}$ denoting the set of learned condition vectors.

Finally, the sequence of step gates is obtained directly from the sequence of $[\mathbf{z}_1,\dots,\mathbf{z}_T]$ such that  $g_t = \sigma(\text{MLP}(\mathbf{z}_t))$.

For ease of notation, we let $\mathcal{P}_{\mathbf{z}} = \textbf{SelectionBottleneck}(\mathbf{z}) = \{ \mathbf{g}, \mathbf{\hat{c}}, \mathbf{\hat{m}}\}$ denote the neural template obtained from the task embedding $\mathbf{z}$. 

\section{Experiments}
In this section, we present the experimental results of applying OC-NMN to different splits of the Arith-MNIST dataset, and compare it against several baseline approaches. 
We analyze the results in Section~\ref{se:results}, and make the following three observations.
(1)~The modular inductive biases employed by OC-NMN alone are not sufficient for our model to achieve systematic generalization to unseen combinations of known concepts.
(2)~However, the same inductive biases can be used to derive a compositional imagination framework that shows promising results in systematic generalization.
(3)~We highlight that 
the ability of the perception model to disentangle object-level attributes is key for generalization to unseen configurations of digit-colors and motivates future work in that direction.

\subsection{Baselines}
Our model can be seen as an adaptation of Neural Module Networks to work on generative visual analogical reasoning tasks. Neural Module Networks are traditionally evaluated in the VQA setting, where a question can be parsed to induce a modular neural network architecture (eg. \emph{neural template}) that acts as an executor. In our setting, the question is not given and the correct architecture 
needs to be inferred from a few input/output support examples. To the best of our knowledge, the only existing neural network model that can readily tackle generative visual reasoning tasks is the Neural Abstract Reasoner (NAR)~\citep{nar}. We compare our model to an object-centric version of NAR (replacing its image encoder with a Slot Attention module and concatenating the slots). We also propose two additional baselines, both having no selection bottleneck: a non-modular baseline, 
where the executor consists of a single GRU cell that takes as input the query slots and the tasks embedding coming from the controller. We denote this baseline \emph{DNC-GRU}.  The second baseline 
consists of a stack of Transformer encoder layers, and takes as input a set composed of the query slots, the controller output, and a CLS token from which we retrieve the final answer. We denote this model \emph{DNC-Transformer}. All architectural details and hyperparameters are described in the Appendix. 

We consider two versions of our model, depending on the number of modules/conditions compared to the ground truth number of concepts 
in the tasks (e.g. conditions and arithmetic operations). 
When the number of modules/conditions is less than the actual primitives needed to solve the training set, we denote our model by~\emph{OC-NMN - less}; otherwise, we  
we denote it by \emph{OC-NMN - enough}.

\paragraph{Text Version.}
In the text version of the benchmark (where each image/output pair is described in natural language), we consider two state-of-the-art language models: FLAN-T5~\citep{chung2022scaling} fine-tuned on our task and GPT-4. 
We evaluate GPT-4 on the easy split and obtain an accuracy of 16 in the best case. More details about the GPT-4 training can be found in the Appendix.

\begin{table*}[t]
\renewcommand{\arraystretch}{1.1}
\centering 
\resizebox{\textwidth}{!}{\begin{tabular}{||c ||  c c c | c c  c |c  c c||} 
 \hline
 Dataset & \multicolumn{3}{c|}{Easy} & \multicolumn{3}{c|}{Medium} & \multicolumn{3}{c|}{Hard} \\ 
 \hline\hline
  Model & Val & Op & Order &  Val & Op & Order & Val & Op & Order  \\
  \hline
 NAR & $97.9_{\pm0.0}$  & $5.9_{\pm 0.7}$ & $33.3_{\pm0.0}$&  $78.4_{\pm2.1}$ & $23.5_{\pm1.8}$ & $31.6_{\pm1.4}$ & $32.9_{\pm3.4}$& $18.5_{\pm0.7}$& $26.6_{\pm2.9}$\\ 
 DNC-GRU & $97.9_{\pm0.8}$ & $5.9_{\pm1.2}$ &$33.5_{\pm0.3}$&  $69.1_{\pm4.3}$ & $21.2_{\pm0.4}$ & $29.6_{\pm0.7}$ & $27.4_{\pm0.5}$& $15.9_{\pm0.5}$ &$23.4_{\pm1.5}$  \\
 DNC-Transformer & $98.3_{\pm0.5}$ & $5.9_{\pm0.4}$ & $33.8_{\pm0.1}$&  $96.9_{\pm 0.8}$ & 25.2 $_{\pm 0.4}$ & 37.8 $_{\pm 1.1}$ & $79.8_{\pm2.1}$& $45.5_{\pm1.8}$ & $65.9_{\pm1.5}$  \\
 OC-NMN - less & $43.1_{\pm33.1}$ & $7.1_{\pm1.1}$ & $15.0_{\pm10.4}$& $66.0_{\pm3.8}$& $22.3_{\pm0.3}$ & $28.7_{\pm1.9}$ & $47.2_{\pm6.5}$& $26.0_{\pm3.9}$ & $43.7_{\pm4.4}$ \\
  OC-NMN - enough & $98.6_{\pm0.5}$ & $5.3_{\pm0.3}$ & $32.8_{\pm0.2}$ & $92.7_{\pm0.8}$& $24.3_{\pm0.7}$& $36.5_{\pm0.5}$ & $82.3_{\pm0.9}$& $41.3_{\pm1.3}$ & $62.3_{\pm2.9}$ \\ 
  \hline \hline
 \multicolumn{10}{||c||}{Text Modality}\\ 
 \hline
 FLAN-T5& 100 & 100 & 99.7&  86.0 & 40.8 & 50.6 & 52.0& 39.0&  47.6\\ 
 \hline

 \end{tabular}}
  \vskip0.2cm
 \caption{Visual reasoning out-of-distribution generalization results. Accuracy averaged over 3 best seeds for the visual models. }\label{table:ood}
\end{table*}
\subsection{Splits}\label{sub : exp}
The different splits we propose and describe in ~\Cref{section:dataset} are designed to evaluate several axes of compositional generalization of both the reasoning and perception components of the models. We want to evaluate: (1)~The ability of the model to generalize to unseen sequences of operations/orders with varying numbers of primitive concepts (e.g., arithmetic operation and condition selection) and samples per task. The hardest split approaches the ARC setting as only a handful of examples per task are seen. Table \ref{table:ood}. (2)~The ability of the perception component to disentangle important factors of variation such as color and digit, which are crucial for solving the tasks at hand (e.g., color represents the number sign). Table \ref{tab:percep}.
We also report the performance of a state-of-the-art language model baseline FLAN-T5 on the text equivalent tasks.
\subsection{Results}
\label{se:results}
\paragraph{Compositional Generalization.}
In  Table \ref{table:ood} we report the performance of the different models on the easy, medium and hard splits. Accuracies are averaged over the 3 best seeds for each model. For each split, we are interested in 2 forms of generalization: performance of the model on a set of different \emph{sequences of operations} (e.g. \emph{Op}) and performance of the model on \emph{digits taken in different order} (e.g. \emph{Order}). The following is a summary of the key observations.  
(1)~Interestingly, all the visual models fail to systematically generalize to both different orders and different operations. It is also the case for OC-NMN, despite the fact that modularity of NMNs have been shown to improve systematic generalization in the context of VQA \citep{andreas2016neural,bahdanau2019closure}. (2)~As expected, OC-NMN fails to learn the tasks at hand when it does not have enough modules and conditions. 
    This is the case where the model has fewer learned modules/conditions than there are primitive concepts (e.g. selection conditions and operations) in the training tasks. (3)~The non-modular DNC-GRU baseline is on par with the others on the easy set, but when the tasks get more complicated (i.e. more concepts required to be learnt and assembled to solve the tasks) it performs subsequently worse than other executors. (4)~DNC-Transformer performs better (in-distribution) than OC-NMN on the medium split but not on the hardest one where only a handful of examples are sampled per task. Note that the DNC-Transformer already contains a form of modularity in the way it processes the input slots (e.g. multi-head attention operations but no selection bottleneck). Overall, OC-NMN and DNC-Transofmer are comparable in performance. (5)~The language model FLAN-T5 finetuned on the same underlying tasks systematically generalizes  OOD when it has enough examples per task. However, it fails to generalize on the hardest split, which is closer to the ARC setting.

\paragraph{Object-level Attribute Disentanglement.}
We evaluate the ability of the perception model to disentangle factors of variation that are relevant to the task (e.g. color and digit, where the color indicates the sign of the digit). We consider two cases that depend on whether the Slot Attention is pretrained on all or only a subset of the digit-color configurations. Once the initial pretraining of the perception module is complete, all the models undergo training on a subset of digit-color configurations and are subsequently tested on held-out configurations.
We report the results of out experiments in Table \ref{tab:percep}. 
The table  illustrates the following observations:
(1) There is a significant decrease in out-of-distribution generalization when the model is pretrained on all versus a subset of the digit-color combinations (even as in-distribution generalization remains high in both cases); and 
(2) when pretrained on all digit-color combinations, the supervised model successfully generalizes to unseen combinations.
Thus, when the perception model is pretrained on all configurations, supervised training on reasoning tasks does not influence generalization to unseen combinations, even if the reasoning module has not been exposed to such combinations during training. 
These results show that the pretraining of the Slot Attention module is key for good generalization to unseen combinations of digit-color pairs. 
This emphasizes the importance of a perception model capable of representing any digit-color configuration during unsupervised pretraining,
as this significantly enhances the generalization capabilities of the model to novel combinations.

\begin{figure*}[t]
    \centering
    \includegraphics[trim={0 0 0 0},clip,width=0.8\linewidth]{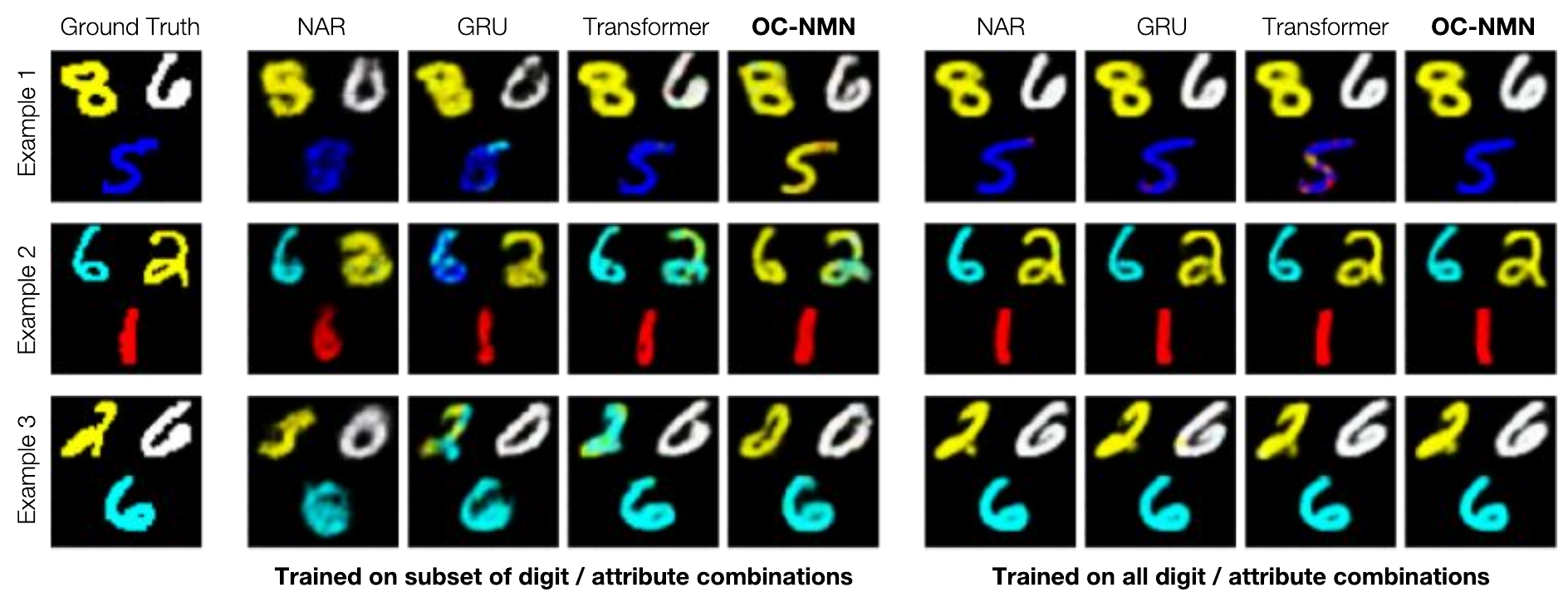}
    \caption{\textbf{Digit-Color disentanglement}. Digits reconstructed from the slots when pretraining Slot Attention on a subset (left) or all of (right) the digit/color combinations.}
    \label{fig : gridpercep}
\end{figure*}

In Figure \ref{fig : gridpercep} we visualize the reconstructed images from the learned slots after training in both pretraining settings. Ground-truth images correspond to digit-color configurations that have been held out during training. In the case where  pretraining is done on all configurations, we notice that training did not influence the ability of the Slot Attention to represent held-out configurations. However, when pretraining is done on the same subset of configurations as training, the slot attention model is not able to represent held-out configurations. It tends to modify the shape and color of digits so that they correspond to configurations it has previously learned to represent.
These findings underscore the need for designing better inductive biases for unsupervised systematic generalization to unseen combinations of object-level attributes and motivate future work for the object-centric community.

\begin{table}[t]
\renewcommand{\arraystretch}{1.1}
\centering
\resizebox{\linewidth}{!}{ \begin{tabular}{||c || c c| c c|} 
 \hline
 Slot Attention Pretraining & \multicolumn{2}{c|}{Pretrain Subset} & \multicolumn{2}{c|}{Pretrain All}\\
 \hline\hline
  Model/Split & Val & OOD Test & Val & OOD Test  \\
  \hline
 NAR & 94.1 & 12.8 & 100 & 96.8 \\ 
 DNC-GRU &  99.3 & 11.6& 100 & 97.4   \\
  DNC-Transformer  & 99.3 & 46.1 & 100& 91.6 \\
 OC-NMN  & 99.6 & 43.6& 100 &   96.5\\ [0.5ex] 
  \hline
 \end{tabular}}
  \vskip0.2cm
 \caption{\textbf{Digit-color disentanglement}. Accuracies on the test set correspond to unseen combinations of digit/color. We distinguish between two settings: unsupervised pretraining on a subset of configurations that are different from the one in the test set (\emph{Pretrain Subset}) and unsupervised pretraining on all the configurations (\emph{Pretrain All}). Accuracies are averaged over the 3 best seeds for each model.}
 \label{tab:percep}
\end{table}

\paragraph{Compositional Imagination.} 
While abstractions, like objects, allow for reasoning and planning beyond direct experience, novel configurations of experienced concepts are possible through imagination. \citet{hoel2021overfitted} takes this notion further by suggesting that dreaming, a form of data augmentation, enhances the generalization and robustness of learned representations. Dreams achieve this by generating new data instances composed of concepts learned or experienced during wake time. 
In this work, we show that the modularity of our model can be used to derive a new paradigm for compositional imagination.

The idea here is that in the same way we select a neural template (sequence of modules, conditions, and gates) using the task embedding output by the controller, we can also sample them at random (from a uniform distribution) to create new imagined scenarios. 
We then train the controller to predict the neural template that generated those imagined samples. The associated loss is called the imagination loss $\mathcal{L}_{\text{aug}} = \mathcal{L}(\mathcal{\hat{P}}_{\mathbf{z}^{\text{im}}}, \mathcal{P}^{\text{im}})$, which can be split into~3 cross-entropies predicting the step gate values, the conditions vector indices, and the processing module indices.
During training, we introduce this loss after a warming period during which the model is trained only on the training data available. We detail the hyperparameters in the Appendix. 

The pseudo-code and training details for this compositional imagination framework are given in the Appendix. 



\begin{figure}[h!]
    \centering
    \includegraphics[trim={10 10 0 0},clip,width=\linewidth]{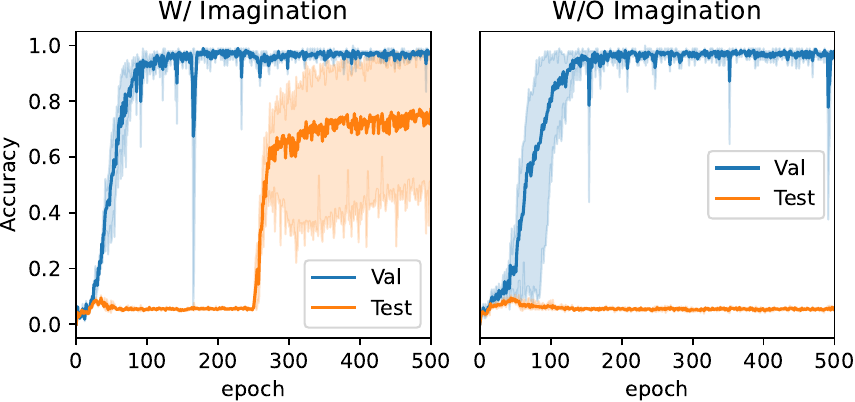}
    \caption{\textbf{Compositional Imagination}. Accuracies in-distribution and out-of-distribution as a function of training epochs. For the model with an imagination component (left), imagination loss is introduced at epoch 250. Results are averaged across the 3 best seeds.}
    \label{fig:imagination_plot}
\end{figure}

In \Cref{fig:imagination_plot}, we show the effect of the imagination loss on the training dynamics, leading to better OOD generalization in the easy split. However, we found that this compositional imagination framework is detrimental in harder splits and suggest that a better prior on how to sample modules/conditions in a non-uniform way is needed. We leave that direction for future work. Additional ablations are given in the Appendix.

\section{Conclusion}
The problem of visual generative reasoning remains challenging for current ML systems~\citep{mitchell2021abstraction,chollet2019on}. In this work, we explore the use of neural module networks~\citep{andreas2016neural} for solving such tasks by leveraging object-centric representations, resulting in our proposed approach called OC-NMN. 
However, we discovered that while NMN's modular inductive biases exhibit strong performance in systematic generalization for visual question answering (VQA) problems~\citep{andreas2016neural,bahdanau2019closure}, such modularity is not enough for 
the visual generative reasoning setting we propose. This disparity can be attributed to the difference in the access to questions in VQA, which provide guidance on assembling the modules into a neural network layout, compared to our setting where the model must infer the appropriate layout based on input-output pairs.
%
To address this challenge, we leverage the generative nature of the problem setup and show how OC-NMN modular inductive biases can be used to  ``imagine'' new unseen problems. 
This involves composing the learned modules in novel ways, applying them to previously seen inputs to create a support set, and training the model to predict back the proper neural network layout.
In encouraging results, we found that this imagination framework helps bridge the generalization gap for easy problems while further research is needed for more complex scenarios.
Additionally, we show that the ability of the perception
to disentangle object-level attributes is key to achieving generalization to unseen digit-color configurations. Our findings underscore the need for designing better inductive biases for object-centric perception models to achieve such disentanglement in an unsupervised manner.

\paragraph{Limitations.} Despite the progress made in our research, it is important to acknowledge the limitations. The benchmarks used in our study are relatively simplistic and do not fully capture real-world complexity. Our method performs well in the easier setting but may struggle with more challenging scenarios. Additionally, the number of hidden ground truth reasoning steps used in our setups remains small (at most 2 operations). These limitations provide opportunities for future research to develop more comprehensive benchmarks, extend the method's applicability, improve complex reasoning capabilities, and ensure practicality in real-world applications.


\bibliographystyle{abbrvnat}
\bibliography{neurips_2023}
\pagebreak
\appendix
\section{Arith-MNIST}
Arith-MNIST is composed of different splits of varying difficulty in terms of the number of primitive concepts that compose the tasks at hand, the number of total tasks, and the number of  examples per task. Each task in the dataset has the same template and consists of a sequence of arithmetic operations applied in a particular order. Each operation and condition to select an argument is sampled from a set of pre-defined concepts. In \Cref{tab:concepts - conditions} and \Cref{tab:concepts - operations} we detail the meaning of all the primitive selection conditions and operations we consider. Since all the tasks share the same template (e.g. ORDER - OPERATIONS), our benchmark is easily extendable to more tasks. 

\paragraph{Splits.} We construct the different splits by selecting: 
a set of primitive concepts and the number of training tasks. The number of training tasks is given by the number of operations-conditions that we consider. For each of the splits, we fix independently a number of operation sequences, and a number of condition sequences. The exact sequences are then sampled at random among all possible sequences. The test splits are  then composed of those held-out sequences. In \Cref{tab : splits - compo} we describe the compositions of the splits we consider in our evaluations. 
For example, in the easy dataset the sequences of operations seen during training are: (\texttt{add}, \texttt{sub}), (\texttt{add}, \texttt{add}),(\texttt{sub}, \texttt{sub}), (\texttt{add}), (\texttt{sub}) and the left-out operation sequence is (\texttt{sub}, \texttt{add}). The training sequence is (\texttt{left}, \texttt{right},\texttt{bottom}).

\begin{table}[h!]
\renewcommand{\arraystretch}{1.1}
\centering
 \resizebox{\linewidth}{!}{\begin{tabular}{|c || c |} 
 \hline
 Conditions &  Meaning \\ 
 \hline\hline
  \texttt{position(X)}&  Select the digit at position X. X can be left, right or bottom \\
  \texttt{max}&  Select the digit with the maximum value. (Default order = left, right, bottom) \\
  \texttt{min}&  Select the digit with the minimum value. (Default order = left, right, bottom)\\ 
 \hline
 \end{tabular}}
 \vskip0.2cm
 \caption{Primitive selection conditions description.}
 \label{tab:concepts - conditions}
\end{table}
\begin{table}[h!]
\renewcommand{\arraystretch}{1.1}
\centering
 \resizebox{\linewidth}{!}{\begin{tabular}{|c || c |} 
 \hline
 Operations &  Meaning \\ 
 \hline\hline
  \texttt{add(X,Y)}&  X + Y. X and Y are the digits values. \\
  \texttt{sub(X,Y)}&  X - Y. X and Y are the digits values. \\
  \texttt{or(X,Y)}& Take the value of (binary(X) \textbf{OR} binary(Y)). \\
  \texttt{xor(X,Y)}& Take the value of  (binary(X) \textbf{XOR} binary(Y)). \\
  \texttt{cat(X,Y)}&  X*10 + Y. X and Y are the digits value. \\
  \texttt{invcat(X,Y)}&  Y*10 + X. X and Y are the digits value.\\ 
 \hline
 \end{tabular}}
 \vskip0.2cm
 \caption{Primitive operations Description.}
 \label{tab:concepts - operations}
\end{table}

\begin{table*}[h!]
\renewcommand{\arraystretch}{1.1}
\centering
\begin{tabular}{||c || c c| c c | c||} 
 \hline
 Dataset & \multicolumn{2}{c|}{Primitives} & \multicolumn{2}{c|} {$\#$ of Training Sequences} & Dataset size  \\ 
 \hline\hline
    & conditions & operations  &  $\#$ of conditions sequences & $\#$ of operations sequences &\\
  \hline
 Easy & \texttt{position} & \texttt{add, sub} & 1  &5 &5000\\ 
 Medium & \texttt{position, max, min} & \texttt{add, sub, or, xor, cat, invcat} & 5&30 &30000 \\
 Hard & \texttt{position, max, min} & \texttt{add, sub, or, xor, cat, invcat} & 55&36 &30000 \\
 Percep & \texttt{position} & \texttt{add, sub} & 1&5 &5000 \\ 
 \hline
 \end{tabular}
 \vskip0.2cm
 \caption{Dataset Splits Composition.}
 \label{tab : splits - compo}
\end{table*}

\section{OC-NMN and Compositional Imagination}\label{sec : im}
Here we describe the training details of the compositional imagination framework. We give in \Cref{compim} the steps of the compositional imagination training. We noticed that good out-of-distribution generalization within this framework was very sensitive to a number of hyperparameters: the most crucial ones were the use of gumbel-softmax in the selection bottleneck, a warm-up phase without imagination, and the learning rate warm-up. We show in \Cref{fig :abla} the effects of the identified important hyperparameters.
 \begin{algorithm}[!h]
 \caption{Compositional Imagination}
 \begin{algorithmic}[1]

 \REQUIRE $\mathbf{X}^{supp}$ \COMMENT{Visual samples seen during training}
 \STATE $\mathbf{S}^{supp} \gets \text{SlotAttention}(\mathbf{X}^{supp})$\COMMENT{Object-centric perception}
 \STATE $\mathcal{P}^{\text{im}} \sim U(\mathbf{g}, \mathbf{C}, \mathbf{M})$ \COMMENT{Sample a neural template}
 \STATE $\mathbf{o}^{\text{im}} \gets \text{Executor}(\textbf{x}^{supp}, \mathcal{P}^{\text{im}})$ \COMMENT{Compute imaginary outputs}
 \STATE $\mathbf{O}^{im}= \{\text{Executor}(\mathbf{x}_i,\mathcal{P}^{\text{im}})\text{ for } \mathbf{x}_i \in \mathbf{X}^{supp}\}$
 \STATE $\mathcal{S}^{\text{im}} \gets \{\mathbf{X}^{supp}, \mathbf{O}^{im}\}$ 
 \STATE $\mathbf{z}^{\text{im}} \gets \text{Controller}(\mathcal{S}^{\text{im}})$
 \STATE $\widehat{\mathcal{P}}_{\mathbf{z}^\text{im}} = \text{SelectionBottleneck}(\mathbf{z}^{\text{im}})$ \COMMENT{Infer sampled neural template}
 \STATE $\mathcal{L}_{\text{aug}} \gets \text{CrossEntropy}(\mathcal{P}, \widehat{\mathcal{P}}_{\mathbf{z}^\text{im}})$

 \end{algorithmic}\label{compim}
\end{algorithm}

 \begin{figure}[b!]
    \centering
    \includegraphics[trim={10 10 0 0},clip,width=\linewidth]{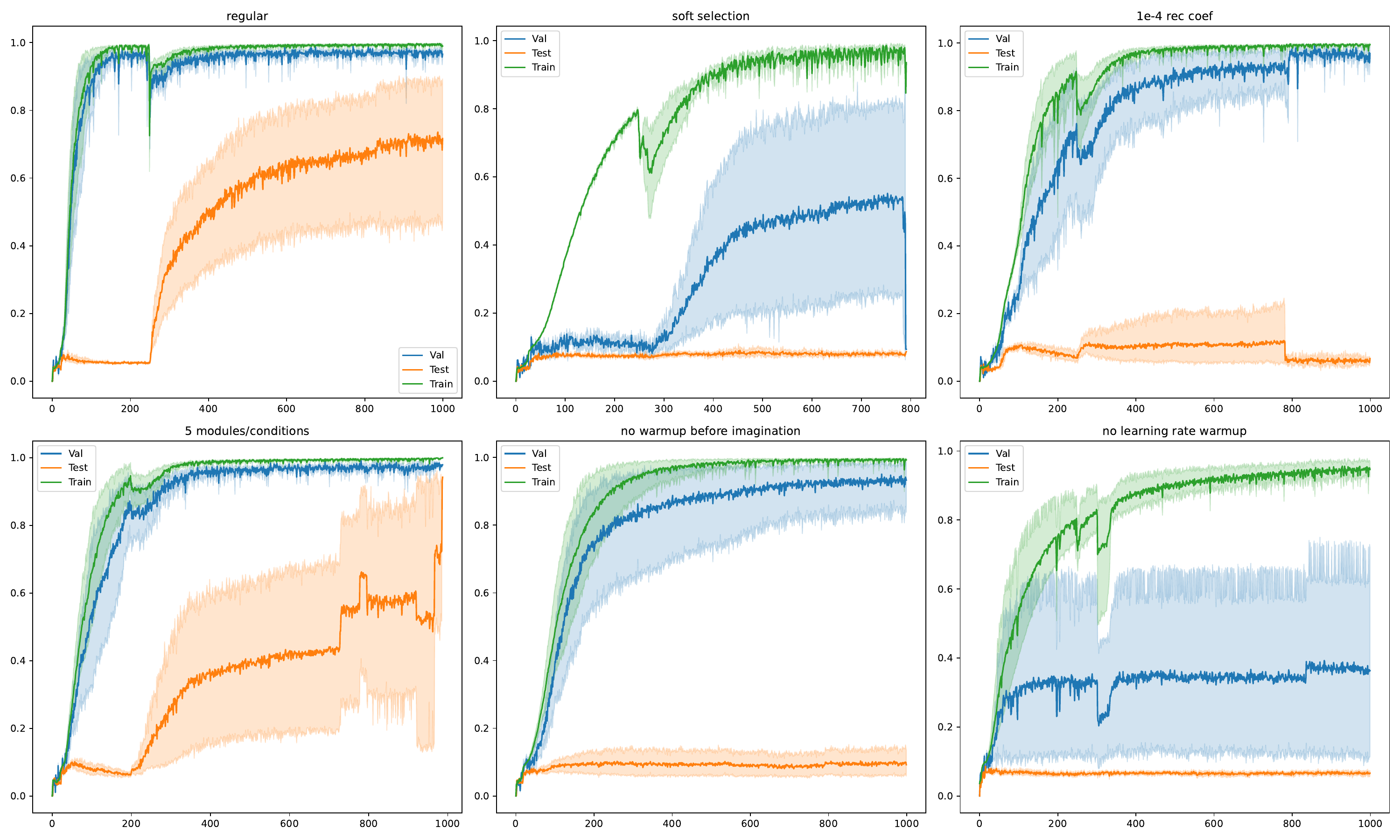}
    \caption{\textbf{Ablation Study}. Training curves  (validation, training, OOD test set) with standard variation shown with shades over 8 seeds. Accuracy is shown as a function of epochs. We consider the following ablations: soft selection instead of gumbel-softmax (top-center), smaller $1e-4$ image reconstruction coefficient (top-right), more modules/conditions (bottom-left), no warm-up period without imagination loss (bottom-center) and no learning rate warm-up (bottom-right). Regular setting is with 3 modules 3 conditions, gumbel-softmax selection and the hyperparameters listed in \Cref{tab : compim hyper}. }
    \label{fig:imagination_curv}
\end{figure}
 \begin{figure*}[h!]
    \centering
    \includegraphics[trim={10 10 0 0},clip,width=\textwidth]{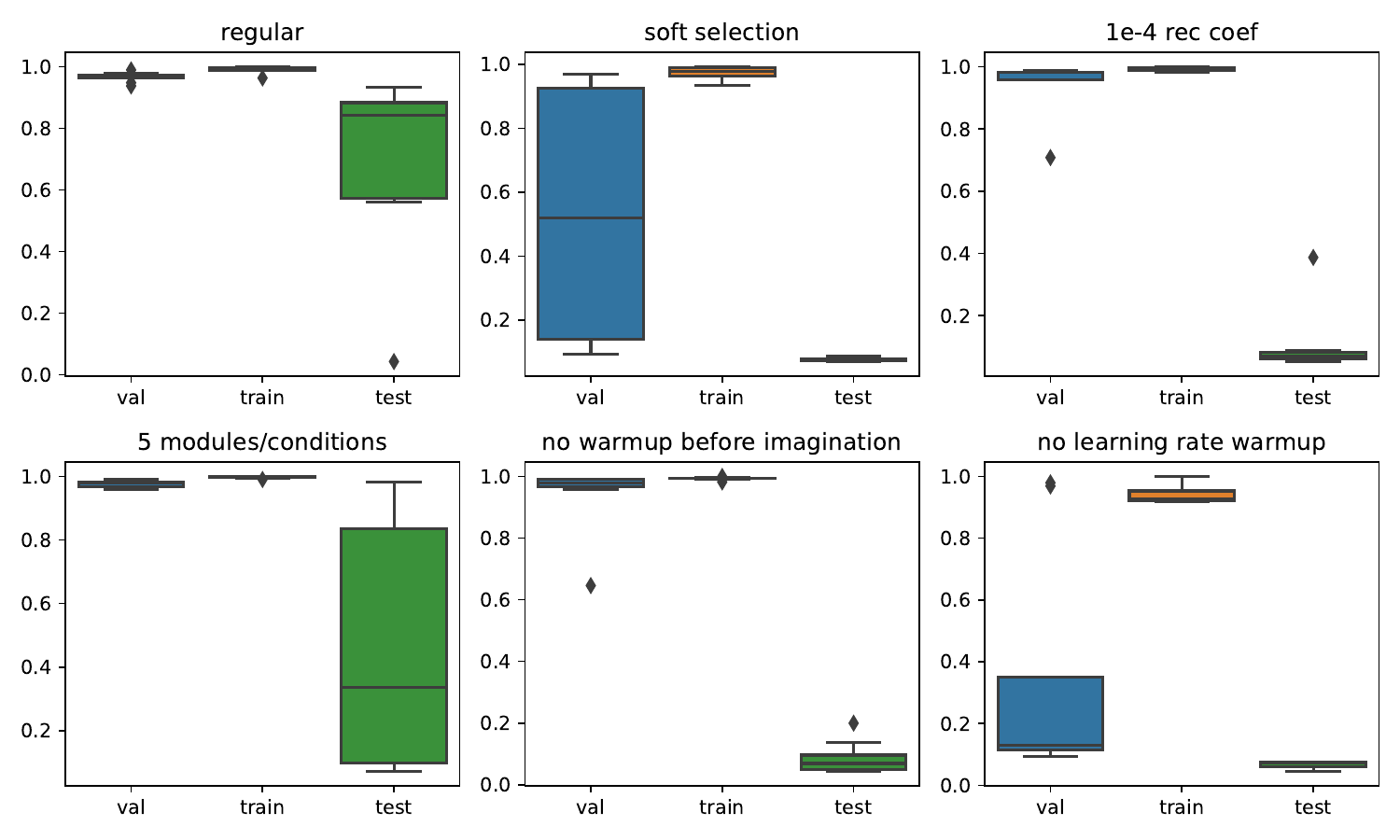}
    \caption{\textbf{Ablation Study}. Final Accuracies for 8 seeds.}
    \label{fig:imagination}
\end{figure*}

\paragraph{Ablation Study.}
We conduct an ablation study of OC-NMN on the easy split to highlight the important hyperparameters of the training that made the compositional imagination framework get good compositional generalization performance. We notice that the use of gumble-softmax and a warm-up period before the imagination loss is introduced is crucial for pushing the selection to be discrete. The coefficient in front of the image reconstruction objective is also important, as it speeds up the training. We also notice  that the learning rate warm-up prevents a collapse mode, in which many seeds simply overfit the training set and would require more regularization. See Figures~\ref{fig:imagination_curv}~and~\ref{fig:imagination}.

\paragraph{Selection Bottleneck.}
The selection bottleneck selects an argument and a module that will update the working memory at each time-step. Here, we give more explanation behind the design of this module, especially the argument selection mechanism. The selection bottleneck uses a condition vector to compare it against the candidate slot arguments with a dot product attention mechanism. However, some conditions (e.g. max, min) require knowledge of other objects in the scene. For that reason, we included an optional stack of 4 self-attention layers before the dot product is applied to the candidate input slots. We use these self-attention layers only for the hard split.

\section{Training and Models Hyperparameters}
In this section, we details of the hyperparameters of the models we considered in our evaluation.

\paragraph{Perception Model.} All the models share the same perception model, which is Slot Attention. For this module, we keep the hyperparameters suggested in the original paper, with latent slot size of $64$, maximum of $4$ slots and different Gaussian parameters per slot to sample the initial guess.

\paragraph{Controller.}
We did not try to optimize the controller architecture. We base our model on the Neural Abstract Reasoner Controller~\citep{nar} and only modify the number of layers of the Differentiable Neural Computer and the perception module to better fit it to an object-centric setting like ours. For all models, we perform hyperparameter search to select the best architecture for each split. 
In \Cref{tab : contro} we detail the hyperparameters selected for the DNC-based controller for all the models and splits.

\begin{table*}[h!]
\renewcommand{\arraystretch}{1.1}
\centering
 \resizebox{\textwidth}{!}
 {\begin{tabular}{|c || c| c |} 
 \hline
 Hyperparameter/Module & Value & Meaning  \\ 
 \hline
$h_{\text{hidden}}$ & 512 & Hidden dimmension of LSTM\\
$l$ & 6& Number of LSTM layers\\
$M$ & $32 \times 64$ & Memory size of DNC \\
$n_r$ & 16 & Number of read heads \\
GRU$_{\text{seq}}$& GRUCell(64,64)& GRU that transforms the last output of the DNC to a sequence of instructions\\

 \hline
 \end{tabular}}
 \vskip0.2cm
 \caption{DNC Controller }\label{tab : contro}
 \end{table*}

\begin{table*}[th!]
\renewcommand{\arraystretch}{1.1}
\centering
 \resizebox{\textwidth}{!}
 {\begin{tabular}{|c || c| c |} 
 \hline
 
 \multicolumn{3}{|c|}{NAR Executor}\\
  \hline\hline
 Hyperparameter/Module & Value & Meaning  \\ 
 \hline
 $L$ & $L = 4$ & Number of layers in executor \\ 
 $N_h$ & $N_h$ = 4 (easy split), $N_h$= 16 ( medium and hard splits) & Number of attention heads in each layer  \\
 $d_i$, $d_o$ & 128, 128 &  input dimension, output dimension of attention layers.   \\
 $d_r$ & 128 & residual MLP hidden dimension\\
 $d_{\text{slot}}$ & 64 & slot dimension \\ 
 $\text{MLP}_{\text{rep}}$ & MLP(256,128) & MLP that transforms input scene representation \\
 $\text{MLP}_{\text{out}}$ & MLP(128, num bits)  followed by sigmoid & MLP that makes the final prediction \\
 num & 11 & Number of binary bits to represent the answer\\ 
 \hline

  \multicolumn{3}{|c|}{DNC-Transformer Executor}\\
  \hline\hline
 $L$ & $L = 6$ & Number of layers in executor \\ 
 $N_h$ & $N_h$ = 6  & Number of attention heads in each layer  \\
 $d_i$, $d_o$ & 128, 128 &  input dimension, output dimension of attention layers.   \\
 $d_r$ & 128 & residual MLP hidden dimension\\
 $d_{\text{slot}}$ & 64 & slot dimension \\ 
 $\text{MLP}_{\text{rep}}$ & MLP(64,128) & MLP that transforms each slot representation \\
 $\text{MLP}_{\text{out}}$ & MLP(128, num bits)  followed by sigmoid & MLP that makes the final prediction \\
 num & 11 & Number of binary bits to represent the answer\\
 \hline
   \multicolumn{3}{|c|}{DNC-GRU Executor}\\
  \hline\hline
 Hyperparameter/module & Value & Meaning  \\ 
 \hline
 $T$ & 4 & Maximum number of updates \\
 $\text{MLP}_{\text{gate}}$  &MLP(128, 1) followed by sigmoid & MLP that predicts the update gate \\
 GRU & GRUCell(256, 256) & GRU Cell that updates the hidden working memory\\
 $d_h$ & 256 & Working memory dimension\\
 
 $d_{\text{slot}}$ & 64 & slot dimension \\ 
 $\text{MLP}_{\text{rep}}$ & MLP(256,128) & MLP that transforms the input scene representation \\
 $\text{MLP}_{\text{out}}$ & MLP(128, num bits)  followed by sigmoid & MLP that makes the final prediction \\
 num & 11 & Number of binary bits to represent the answer\\
 \hline
   \multicolumn{3}{|c|}{OC-NMN Executor}\\
  \hline\hline
 Hyperparameter & Value & Meaning  \\ 
 \hline
  $T$ & 2 (easy) 4 (medium, hard splits) & Maximum number of updates \\
  $\text{MLP}_{\text{rep}}$ & MLP(64,64) & MLP that transforms each slot representation to be used by the executor\\
  GRU module & GRUCell(64,64) & Module in the learnt module library \\
  $\mathbf{C}$ & Param($N_c, 64$)& Learnt conditions queries for the argument selection.\\
  MLP query & MLP(64,64) & MLP to extract keys from the slot for the argument selection.\\
  $\text{MLP}_{\text{gate}}$  &MLP(64, 1) followed by sigmoid & MLP that predicts the update gate \\
 $\text{MLP}_{\text{out}}$ & MLP(64, num bits)  followed by sigmoid & MLP that makes the final prediction \\
 num & 11 & Number of binary bits to represent the answer\\
 \hline
 \end{tabular}}
 \vskip0.2cm
 \caption{Executor modules hyperparameters. }
\label{tab : executor}
\end{table*}
\subsection{Executor Modules}
We describe here the different executor modules along with their hyperparameters. All the hyperparameter values are listed in~\Cref{tab : executor}. 
\paragraph{NAR.}
The NAR executor is composed of a stack of layers composed of a self attention and a cross-attention step. The self-attention is performed on the set of support and query input representations. Let  $[\mathbf{s}_1^{supp}, \dots, \mathbf{s}_{N_s}^{supp},\mathbf{s}_1^{squery} ]$ denote the concatenation of the support and query input representation (e.g. concatenated object-centric slots for each scene representation), $\mathbf{z}$ the output from the controller, and $\text{MHA}(q, k, v)$ the multi head dot product attention operation over queries $q$, keys $k$, and values $v$ followed by a residual MLP layer. The output  $\mathbf{o}$ of a layer in the NAR executor corresponds to: 

\begin{align}
\mathbf{h} &= [\mathbf{s}_1^{supp}, ..\mathbf{s}_{N_s}^{supp},\mathbf{s}_1^{squery} ] \\
\mathbf{h} &= \text{MHA}(\mathbf{h}, \mathbf{h}, \mathbf{h}) \\
\mathbf{q} &= [\mathbf{z},\mathbf{h}]\\
\mathbf{o} &= \text{MHA}(\mathbf{q}, \mathbf{h}, \mathbf{h})
\end{align}
 The NAR executor consists of a stack of a  $L$ layers described above. The final output is taken from the token that corresponds to the query position. 

 \paragraph{DNC-GRU.}
The DNC-GRU executor corresponds to the non-modular version of our OC-NMN. It does not have any selection bottleneck and is composed of a single GRU cell that performs a sequential update of working memory from which the final answer is retrieved. Let $\mathbf{h}_t$ denote the working memory at time step $t$,  $\mathbf{z} = [\mathbf{z}_1, \dots, \mathbf{z}_T]$ the output from the controller and $\mathbf{s}^{\text{query}}$ the query input representations. The working memory at time-step $t$ is updated as follows: 
\begin{align}
    \mathbf{h}_t = \alpha_t \cdot\text{GRU}(\mathbf{h}_{t-1}, [\mathbf{s}^{\text{query}}, \mathbf{z}_t]) + (1-\alpha)\cdot\mathbf{h}_{t-1}
\end{align}
where $\alpha_t = \sigma(\text{MLP}(\mathbf{z}_t))$ corresponds to an update gate that controls the number of updates performed to the working memory in a soft way. 

 \paragraph{DNC-Transformer.}
Let $\mathbf{s}^{\text{query}} = [\mathbf{s}^{\text{query}}_1, \mathbf{s}^{\text{query}}_N]$ denote the set of $N$ object-centric slot representation of the input query and $\mathbf{z}$ the output from the controller. The DNC-Transformer executor consists of a stack of transformer layers that is performed over the set composed of the inputs slots, a \emph{CLS} token, and the output of the controller $\mathbf{z}$. The final result is retrieved from the CLS token at the end of stack.

\subsection{Training}
\paragraph{Unsupervised pretraining.}
All models are trained with the same Slot Attention perception module. This perception module is pretrained with the same hyperparameters described in the original Slot Attention article \citep{locatello2020objectcentric} with a maximum of $N_s = 4$ slots and different learned Gaussian parameter per slot. In our setting, we found that at the end of the pretraining phase, each slot corresponds to a fixed position in the image (left, right, or bottom). 

\paragraph{Supervised training.}
After pretraining, the slot attention module weights are loaded and trained along with the rest of the model. In order to keep the object-centric property for the remainder of the training, we add the same reconstruction loss that was used for the pretraining weighted by $\alpha_{\text{rec}} = 2e^{-4}$. All the models are trained with an Adam optimizer \citep{kingma2014adam} with a learning rate of $l_r = 3e^{-4}$ and a batch size of $32$. In the compositional imagination experiments, we found that using a learning rate warm-up of ~20 epochs helps the model not to collapse.

\paragraph{OC-NMN optimization hyperparameters.} As shown by the ablations in \Cref{sec : im}, OC-NMN is sensitive to a couple of hyperpermaters in the compositional imagination setting. We list them in \Cref{tab : compim hyper}.
\begin{table}[h!]
\renewcommand{\arraystretch}{1.1}
\centering
 \resizebox{\linewidth}{!}
 {\begin{tabular}{|c || c| c |} 
 \hline
 Hyperparameter & Value & Meaning  \\ 
 \hline
$T$ & 250 & Number of epochs without imagination loss\\
$w$ & 20 & Number of epochs of linear learning rate warm-up\\
$\alpha_{im}$ & 50 & Imagination coefficient \\
$\alpha_{rec}$ & $2e^{-4}$ & Image reconstruction coefficient\\
$t$ & 3 & Gumbel-Softmax Temperature\\
lr & $3e^{-4}$ & Learning rate \\
 \hline
 \end{tabular}}
 \vskip0.2cm
 \caption{Compositional Imagination Hyperparameters}
 \label{tab : compim hyper}
 \end{table}

\begin{figure*}[b!]
    \centering
    \includegraphics[width=0.8\linewidth]{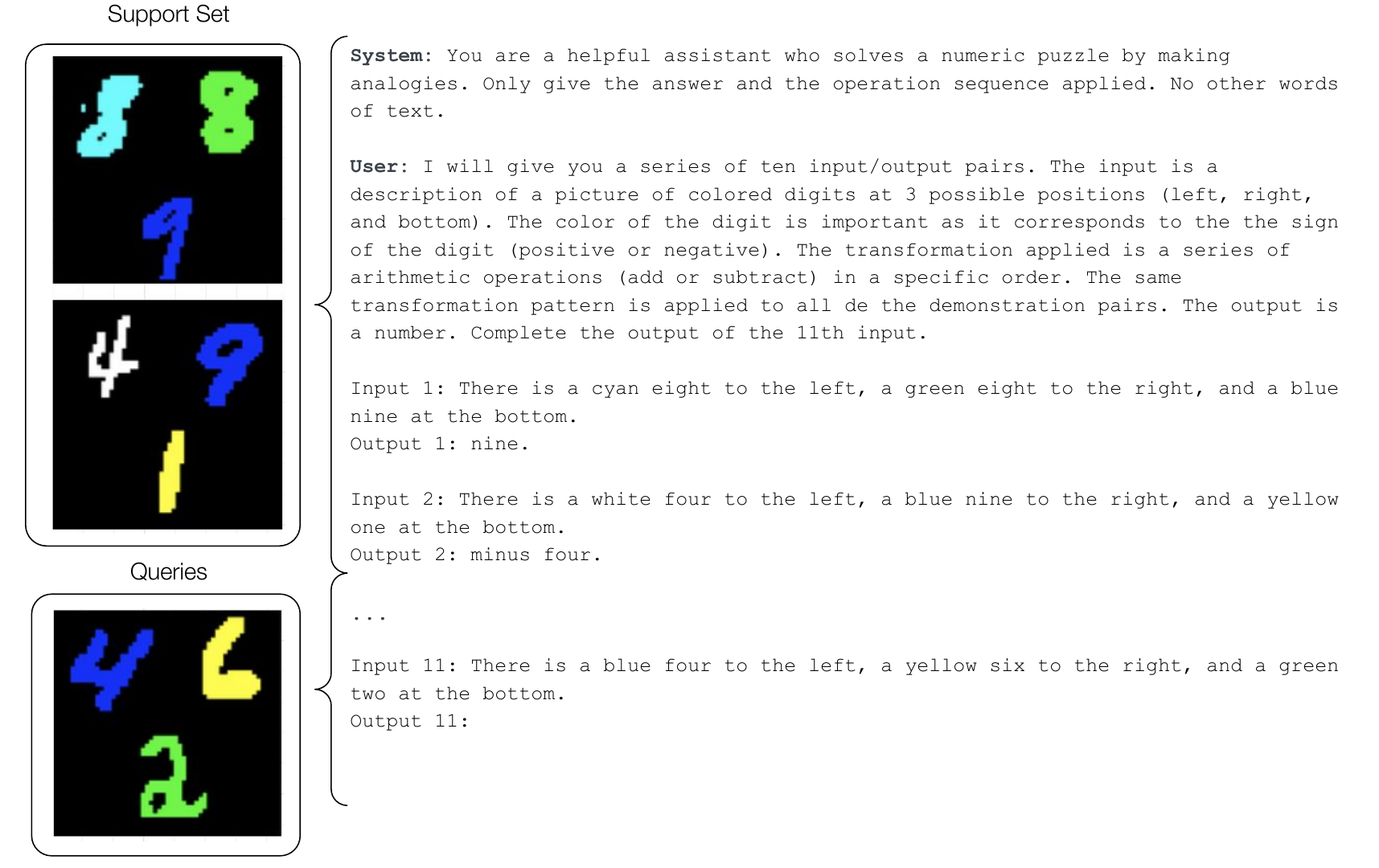}
    \caption{Setup of the GPT-4 experiments. The visual demonstrations are transformed into textual descriptions and fed to the model. GPT-4 is asked to generate the response and the operation sequence used to find the result for each task.}
    \label{fig:gpt-4}
\end{figure*}
\section{GPT-4 Experiments}
Inspired by the growing interest in large language models and their analogical reasoning capabilities~\cite{webb2022emergent}, 
we test GPT-4 on a subset of the tasks in Arith-MNIST. 
GPT-4 is a multi-modal AI system created by OpenAI and  is known to have zero-shot reasoning capabilities on a wide range of tasks~\cite{webb2020emergent}. 
Since we test on the language-only version of GPT-4, we convert the visual input of Arith-MNIST into a textual description of what is seen in the image.
We then use the API provided by OpenAI to prompt GPT-4 using its ``system'' and ``user'' components to introduce  the problem setup and pass on the individual task descriptions.  
We use a prompt structure similar to ~\citep{moskvichev2023conceptarc}, although we give the model some more context and information about the problem. 
We test GPT-4 on 100 samples from the easy split. 
See Figure~\ref{fig:gpt-4}. 

The accuracy of GPT-4 on the 100 easy samples from Arith-MNIST is 11\% when we consider only the output number (ignoring the predicted sequence of operations).  
In only 4\% of cases, GPT-4 could correctly predict the  sequence of operations that would generate the correct answer. 

In an attempt to improve GPT-4 performance, we experimented with different prompts having different levels of granularity: giving one full example in the initial prompt, providing a step-by-step guide to the solution, asking GPT-4 to explain its reasoning for each sample solution. The best test result we obtained was 16\% on the same test samples.

\section{Comparison with Neural Module Networks}
 \begin{figure*}[h!]
    \centering
    \includegraphics[trim={10 10 0 0},clip,width=\textwidth]{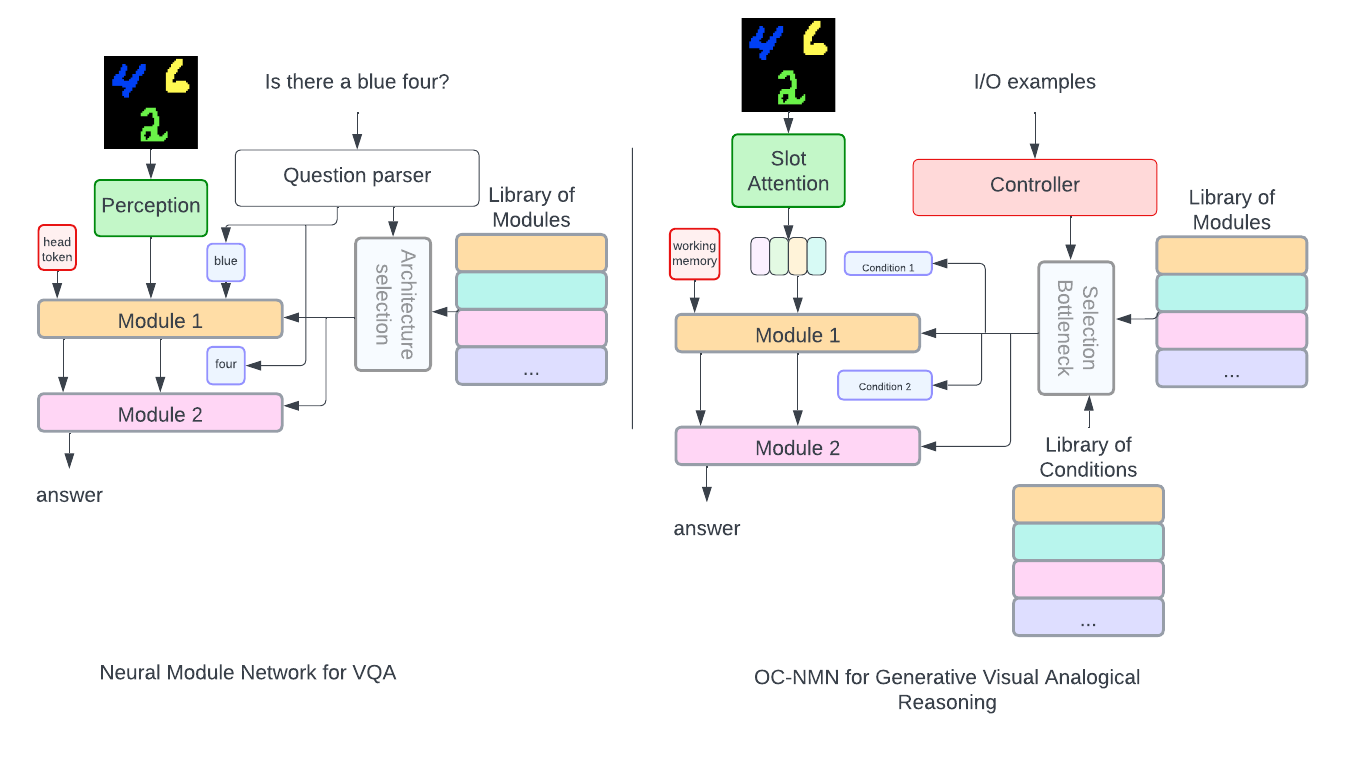}
    \caption{Comparison of Neural Module Networks in the context of VQA (left) and OC-NMN in the context of generative analogical visual reasoning (right)}
    \label{fig:nmn vs}
\end{figure*}
\end{document}